# NusaAksara: A Multimodal and Multilingual Benchmark for Preserving Indonesian Indigenous Scripts

**Muhammad Farid Adilazuarda**[1], **Musa Izzanardi Wijanarko**[2], **Lucky Susanto**[2], **Khumaisa Nur'aini**[2], **Derry Wijaya**[2], **Alham Fikri Aji**[1]

[1]MBZUAI [2]Monash University Indonesia

## Abstract

Indonesia boasts over 700 languages, with a rich diversity of writing systems. However, most NLP development has been based on romanized text, with limited support for native writing systems. We present NusaAksara, a novel public benchmark for Indonesian languages that includes their original scripts. Our benchmark covers both text and image modalities and encompasses diverse tasks such as image segmentation, OCR, transliteration, translation, and language identification. Our data is constructed by human experts through rigorous steps. NusaAksaracovers 8 scripts across 7 languages, including low-resource languages not commonly seen in NLP benchmarks.Among the scripts covered in this dataset, the Lampung script is included despite being unsupported by Unicode. We benchmark our data across several models, from LLMs and VLMs such as GPT-4o, Llama 3.2, and Aya 23 to task-specific systems such as PP-OCR and LangID. Our results reveal that most NLP technologies struggle with Indonesias local scripts, with many achieving near-zero performance.[1]

## 1 Introduction

*"The death of a language is the loss of its knowledge." - Hywel Coleman*

Indonesia is home to a remarkably diverse range of more than 700 languages (Aji et al., 2022), many of which were originally written in their own scripts. However, in recent times, speakers have increasingly adopted romanized scripts, leading to the gradual decline (Fogg, 2015) and neglect of these traditional writing systems (Matthews, 1983; Ibrahim, 2011). Consequently, Indonesian-specific NLP technologies, like other multilingual low-resource technologies, overlook local scripts (Kirmizialtin and Wrisley, 2020; Khan et al.), reinforcing a cycle that further diminishes their use. These local writing systems, locally known as *aksara*[2], are not just tools for communication but also vessels of cultural identity (Taylor, 1998; Adilazuarda et al., 2024b) and repositories of historical knowledge (Florida, 1995). Although in Indonesia, Bahasa Indonesia serves as the *lingua franca*—uniting the country's diverse linguistic communities, revitalizing local languages remain vital to national identities and cultural heritage (Suhendi, 2025).

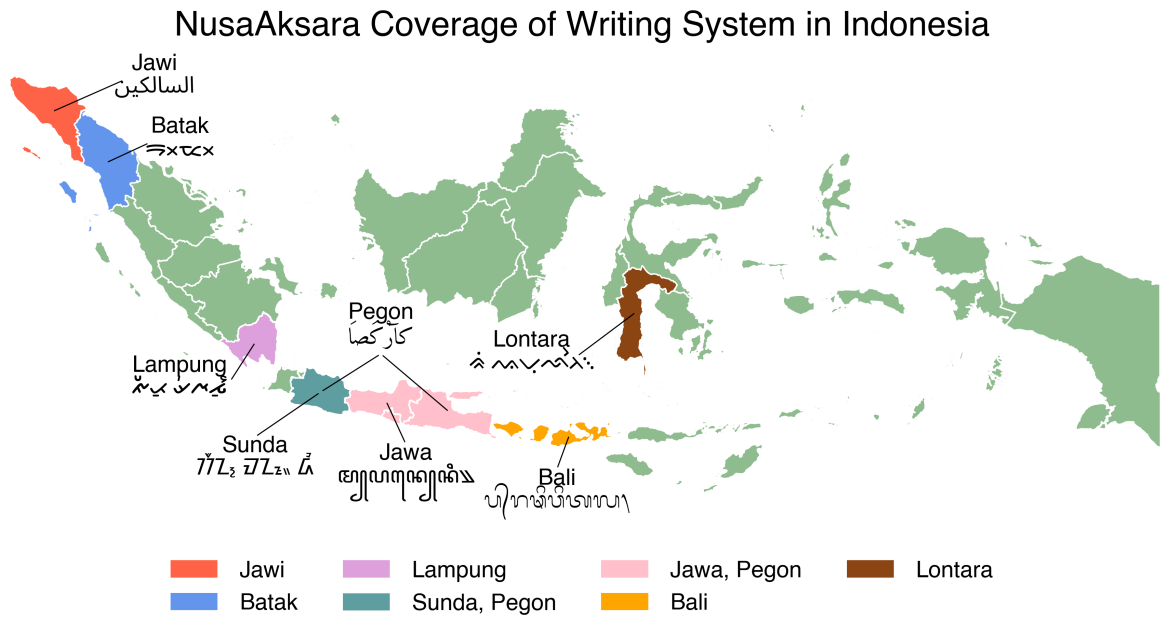

Figure 1: NusaAksara benchmark script coverage.

In this paper, we investigate NLP data for Indonesian languages, which is predominantly collected in romanized form[3]. Supported by previous research (Adilazuarda et al., 2024a), we also find that most models barely recognize the traditional scripts. The scarcity of documented resources, combined with the lack of technological support, poses significant challenges to their preservation (Perdana, 2024). To address this gap, we develop NusaAksara—a comprehensive benchmark and define key tasks that leverage NLP techniques to safeguard and revitalize Indonesia's traditional scripts. Our dataset includes scanned documents written in 8 different scripts. Through expert annotation and validation, we transcribe, transliterate, and translate (into In-donesian) the data. This dataset can be used for a variety of tasks across different modalities, including segmentation, optical character recognition (OCR), transliteration, translation, and language identification (LID).

[1]We release our benchmark dataset in huggingface https://huggingface.co/datasets/NusaAksara/NusaAksara.

[2]The word *aksara* originates from Sanskrit and now means the letters or basic symbols used in a writing system of a language—in other words, *script*.

[3]Throughout this discussion, we define *romanized* as referring to the Latin script, and *local aksara* as referring to the original local script.

Despite claims of multilingual capability (Qin et al., 2024; Huang et al., 2024; Adilazuarda et al., 2022; Choudhury and Deshpande, 2021), many LLMs and other models, including those specifically designed for Indonesian languages, struggle with our benchmark. Opaque models like GPT-4 and Gemini yield some decent results, but there remains significant room for improvement.

In summary, our contributions are as follow:

- We introduce NusaAksara, a novel conservation project focused on local scripts in Indonesia.
- Our dataset covers 8 distinct local scripts and 7 languages. Most of the languages are considered low-resource, and one of the scripts does not have a registered Unicode.
- We define several tasks for this dataset, including image segmentation, OCR, transliteration, translation, and LID.
- We analyze current NLP data and models in terms of Indonesian script coverage, demonstrating their shortcomings.
- We benchmarked NLP models and methods, ranging from LLMs such as GPT-4 to specific methods such as NLLB for translation, revealing their underperformance for this task.

# 2 Indonesian NLP Resources in Local Scripts

## 2.1 Part 1: Data Study

With over 700 languages spoken in Indonesia, only a few are documented in NLP datasets, whether for pretraining, fine-tuning, or benchmarking purposes. Recently, there has been an encouraging increase in efforts to build resources for Indonesian NLP. However, the vast majority of these resources are written in Latin script, rather than in their original scripts. In this section, we examine the current state of available data with respect to their written scripts.

| Model | Dataset(s) | ID Native Scripts |
|---|---|---|
| NLLB-3.3B | CC, OSCAR, Paracrawl, CCNet | 0.0% |
| bloomz-7b1 | ROOTS, CC, MC4 | 0.0% |
| Cendol-MT5 | Cendol | 0.015% |
| Llama-3.1-8B | Mixed Web | 0.0% |
| Llama3.2-11B | MultiModal Web | 0.0% |
| Sailor-7B | SlimPajama, SkyPile, MADLAD-400, CC100 | 0.018% |
| aya-23-8B | Aya Collection | 0.0% |

Table 1: The distribution of scripts within the model serves as a proxy for the corresponding dataset, illustrating the frequency of unique tokens associated with native Indonesian (ID) scripts, including the cumulative proportions of aksara Jawa, Sunda, Lontara, Bali, Rejang, and other related scripts.

**Lack of Representation in LLM** LLMs are primarily trained on massive multilingual datasets, such as PILE (Gao et al., 2020), OSCAR (Ortiz Suárez et al., 2020), CommonCrawl, and Aya, which offer vast linguistic diversity. However, despite supporting numerous languages, these datasets are heavily skewed toward Latin-based scripts, even for languages that traditionally use other writing systems.

To better understand this disparity, we analyzed script distributions across various language models by comparing the prevalence token of Latin-derived scripts against that of indigenous or historical scripts. We extracted tokens from pretrained models and utilized the `unicodedata`[4] to map them to their respective scripts (Appendix B) .

Despite extensive multilingual capabilities of LLMs, the representation of Indonesian local scripts across various relevant datasets remains extremely low or even entirely absent, as shown in Table 1. While models like CENDOL-MT5 (Cahyawijaya et al., 2024a) and Sailor-7B (Dou et al., 2024) exhibit a slightly improved representation of local scripts owing to their more diverse datasets tailored for Indonesian and South-East Asian languages, they still do not achieve an equitable representation. This imbalance constrains the linguistic richness that models can capture and disproportionately affects traditional scripts, resulting in decreased representation within multilingual

[4] `unicodedata` module is a Python library for accessing Unicode character properties. See: https://docs.python.org/3/library/unicodedata.html

models (Adilazuarda et al., 2024a).

**Lack of Representation in Downstream Benchmark** Labeled or benchmark data is equally important in the modern NLP landscape. SEACrowd (Lovenia et al., 2024) is a recent crowd movement that gathers NLP datasets for South-East Asian languages, respectively, and managed to gather 502 datasets, 105 of which contain Indonesian regional languages. Unsurprisingly, the majority of them are written in romanized scripts. Specifically, we found only two datasets that explicitly claim to be written in local scripts, namely AMADI_LontarSet (Kesiman et al., 2016) and DeepLontar (Siahaan et al., 2022).

### 2.2 Part 2: Non-NLP Resources

Before Dutch colonization, many Indonesian languages had their own indigenous scripts that were used for literature, government documents, and religious texts in Indonesian Hindu-Buddhist kingdoms (e.g. Majapahit) and later in Indonesian Islamic kingdoms (e.g. Mataram). However, during the colonial era, similar to other parts of the world where colonial codification took place (Yelle, 2012; St-Pierre, 2000), Romanized standard orthography was enforced, which results in marginalization of indigenous scripts in Indonesia. The change from native to Latin script means that some sounds or meanings are lost. For example, there are different ⟨e⟩ sounds in Javanese native characters such as ꦼ and ꦺ that are lost when transliterated as the character ⟨e⟩ in the current Indonesian Enhanced Spelling System (EYD) that continues this colonial policy after independence.

Due to the lack of support for traditional scripts, such as proper keyboards or even supported Unicode standards, most speakers resort to romanizing their writings in digital contexts, including social media and online messaging. Younger generations can no longer read historical texts or pre-colonial literature, which results in cultural loss and displacement as future generations lose access to centuries of traditional knowledge, literature, and history and see their own past as foreign (Cummings, 2002).

However, it is crucial to explore non-typical NLP contexts where local scripts continue to hold significance. These scripts remain integral to everyday life and appear in historical artifacts, cultural expressions, and educational materials. Here, we provide examples to illustrate why preserving these scripts matters:

**Educational Purposes** Local scripts are part of the curriculum in Indonesian schools, where students are taught the basics of reading and writing these scripts as a way to connect with their heritage, strengthen linguistic diversity, and help prevent language extinction.

**Street Signs and Public Use** In certain regions, local scripts are still used on street signs such as in Yogyakarta and Bali.

**Historical Manuscripts** Local scripts are often found in ancient manuscripts that hold invaluable historical, scientific, and cultural knowledge. For instance, palm-leaf/lontar manuscripts written in Balinese script offer insights into traditional medicine, astrology, and historical events. Losing these scripts would mean losing access to this reservoir of knowledge.

**Historical Legal Documents** Documents such as land deeds, loan agreements, and family records from earlier times were often written in local scripts. These documents are not only important for historical research but also occasionally for legal and familial purposes today. Preserving the knowledge of these scripts ensures that these records remain accessible and interpretable.

## 3 Corpus Construction for Local Scripts

### 3.1 Script of Focus

We focus on eight Indonesian scripts and the languages they traditionally represent, as shown in Table 2. In addition to proposing a new dataset in these local scripts, which are rarely found in typical Indonesian datasets. We also cover low-resource languages that are often absent from multilingual benchmarks. More details on each script and its corresponding language can be found in Appendix A.

### 3.2 Dataset Creation

#### 3.2.1 Source

**Resource Digitization** Our dataset is compiled from a variety of sources, including historical manuscripts, literary works, books, religious texts, magazines, and educational literature. These resources provide authentic examples of language use in local scripts. We carefully selected sources that represent the linguistic and cultural richness of each language to cover a diverse range of topics and styles despite the lack of digital media containing local Indonesian scripts.

| Script | Lang | Original Source #books | #pages | Content type | Final Resulting Data #sents | #char | Example |
|---|---|---|---|---|---|---|---|
| Lampung† | ljp | 4 | 608 | Local books | 1,029 | 7,959 | [illegible] |
| Jawi | zsm | 9 | 838 | Classical Malay documents | 1,018 | 19,712 | السالكين |
| Bali | ban, kaw⋆ | 3 | 518 | Religious texts | 459 | 22,179 | [illegible] |
| Batak | bbc, btx⋆, btm⋆ | 2 | 294 | Traditional manuscripts | 847 | 6,357 | [illegible] |
| Jawa | jav | 39 | 2271 | Historical Texts, Community Contributions | 816 | 22,560 | [illegible] |
| Lontara | bug | 5 | 362 | Traditional manuscripts | 477 | 11,945 | [illegible] |
| Pegon | jav | 6 | 1292 | Historical & religious texts | 964 | 23,249 | اصَكْراكَ |
| Sunda | sun | 7 | 954 | West Java archives | 823 | 14,085 | [illegible] |

Table 2: Data statistics and examples of our data. †The Lampung script is written with a custom font, as there is no proper Unicode support otherwise. ⋆We were unable to obtain sufficient data for these languages; therefore, they have been excluded from the final benchmark dataset.

Initially, we planned to gather data from the National Library of Indonesia. However, after our visit, we faced two major challenges: the limited availability of recent textbooks written in local Indonesian scripts and the strict policy that allows only 10 pages to be scanned per day. We then sourced books from online marketplaces, purchasing 2-9 books for each identified script. This process took several weeks until all physical books were delivered. We also obtained additional Javanese script resources from old local magazines in one of our authors' personal collection. Moreover, we received a digitalized e-book from local communities as supplementary material for Javanese script. Next, we manually unbound (see appendix E, Fig. 6) and scanned all 75 books totaling of 7,137 pages for digitization (see Table 2).

**Data Processing** Since the digitized books still contained significant romanized text, we developed a system to detect local scripts in the digitized resources. We fine-tuned PaddleOCR (Du et al., 2020) detection model to recognize local scripts in our data while ignoring the Latin script. To train the model, we hired two annotators to create labeled bounding boxes distinguishing local scripts from Latin (see Appendix E for example). They annotated 100 pages for each script, after which we trained a DB-based text detection model. While the resulting model isn't flawless, it significantly speeds up the subsequent human annotation processes (Section 3.2.3).

We sampled and extracted no more than 10% of the content of each book across random chunk of text, compiling approximately 1,000 segmented images per local script to be transcribed, transliterated and translated by native speakers. We release our data under non-commercial license.

#### 3.2.2 Annotators Hiring

To annotate the dataset, we collaborated with native speakers, educators, linguists, and members of the grassroots community who are actively involved in the preservation of local scripts. In particular, we engaged with the Aksara di Nusantara community[5], a group that preserves various local-script initiatives in Indonesia. We also conducted several discussions with local grassroots communities. To get our pool of annotators, we announced an open call for annotators, then asked them to complete a short pre-test.

The test assessed three key competencies: **1. Transcription:** Typing and transcribing text in local script. **2. Transliteration:** Converting text from the local script to Latin script. **3. Translation:** Translating the text into Indonesian.

Out of 88 respondents, we selected one annotator per script based on their performance in these competencies and their proven familiarity with both the script and its corresponding language. We do this approach to ensure accurate and culturally sensitive annotations. We also conducted a follow-up validation phase with another pool of selected annotators to clarify ambiguity in the text and to maintain consistent annotation guidelines.

#### 3.2.3 Data Annotation

Our annotation process is conducted using Label-Studio[6]. Before starting the annotation process, we train our annotators with a pre-recorded video tutorial of the annotation process. We then set up a Zoom call with the annotators to provide additional training and share the annotation guidelines (Appendix I). The annotators are instructed to:

[5] https://aksaradinusantara.com
[6] https://labelstud.io/

1. Fix the bounding box of the local script inferred by our fine-tuned PaddleOCR system previously discussed in Section 3.2.1.
2. Digitize the text in the bounding box by writing it in the respective local script.
3. Transliterate the text into romanized script.
4. Lastly, translate the text to Indonesian.

The data annotation steps are illustrated in Figure 2, and the annotation interface is shown in Figure 5 in the Appendix.

#### 3.2.4 Data Validation

After annotation, a human validation step ensured data quality. Appendix H details the validation process, computing the agreement between the annotator and the corresponding validator for the transcription, transliteration, and translation tasks.

In general, both transcription and transliteration achieved low character and word error rates (i.e., CER and WER), indicating a high level of agreement. Most revisions focused on standardizing spelling variations, ensuring correct transcription of scripts, and improving phonetic accuracy. However, the transliteration of Lontara demonstrated higher CER and WER scores (0.0619 and 0.2137) due to standardization challenges with the representation of final consonants in Latin (e.g., *lontarak, lontaraq, lontara*). Jawa script also displayed variations in the phonetic representation of characters in Latin (e.g., *dha/da*), inconsistencies in capitalization, and instances of missing double letters in compound words (e.g., *harapane* instead of *harapanne*).

The overall translation agreement was high across all scripts, with BLEU and chrF++ scores exceeding 90. However, Lontara recorded the lowest scores, 48.92 for BLEU and 66.07 for chrF++, mainly due to paraphrasing. For instance, the Lontara annotator translated a script to *"Yang mulia dan dahi"* (The noble and the forehead). The validator translated the script to *"menampakkan kemuliaan terutama dahi"* (Displaying nobility, especially on the forehead), resulting in a sentence that is more natural and fluent in Indonesian.

### 3.3 The Curious Cases of Preserving Local Scripts

**Aksara Lampung, the non-unicode script** The Lampung script presents a unique challenge, as it has not yet been officially recognized or standardized in the Unicode system. Consequently, digital preservation of this script become significantly more difficult. For instance, we required the annotator to write the annotation in a separate document rather than in our own Label Studio platform as it needs a specialized font to display Lampung text correctly.

**One Script, Two Languages** Some local scripts can represent more than one language, which adds another layer of complexity to our preservation effort. For instance, the Batak script is used by both Batak Karo (btx) and Batak Toba (bbc), while the Lontara script represents Bugis (bug) and Makassarese (mak). Additionally, Pegon (and Jawi, respectively) are employed for writing Javanese (jav) (and Malay (zsm) resp.), and Arabic. These overlaps pose interesting questions for data annotation and corpus building, as multiple language communities need to coordinate standardization efforts, develop orthographic conventions, and create NLP resources that accurately reflect each language.

### 3.4 Task Formulation

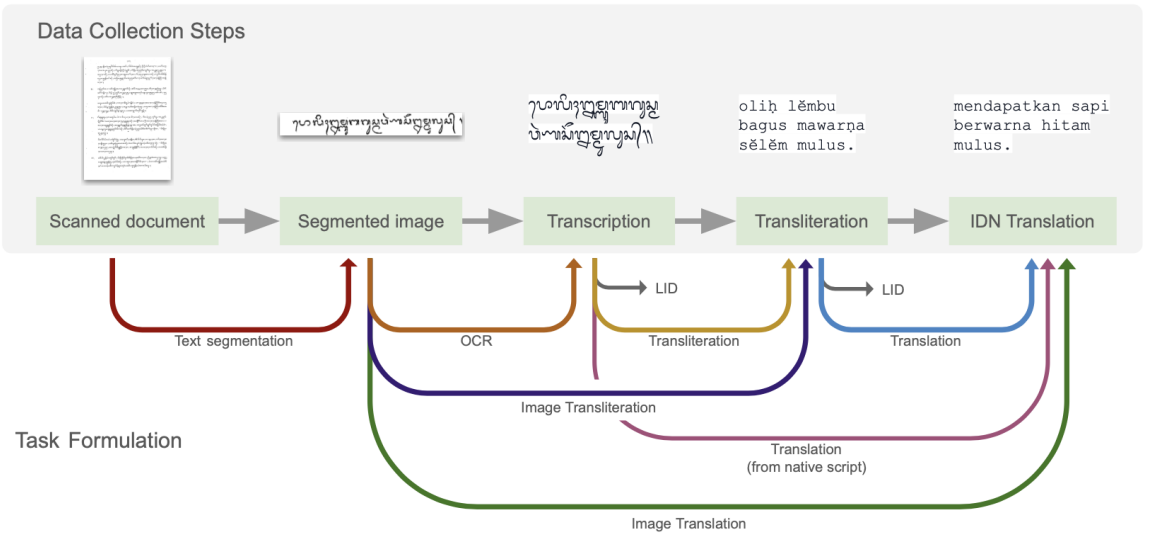

Figure 2: Task formulation pipeline

From our data annotation pipeline, we gathered data across various formats and modalities, starting from scanned documents, segmented text data, transcription, transliteration, and Indonesian translation. This allows us to construct nine distinct tasks to benchmark models on our data, as illustrated in Figure 2.

**Text Segmentation** Extracting script bounding boxes from images of scanned documents.

**OCR** Converting text segment images into machine-readable local scripts.

**Transliteration** Converting text from local scripts into romanized forms.

**Image Transliteration** Transliterating segmented text images directly into romanized text.

**Translation** Translating text into Indonesian, with two formats: one from romanized scripts and another from original scripts.

| Task | Metric | Opaque Models | | Vision Models | | Language Models | | Specific Systems | |
|---|---|---|---|---|---|---|---|---|---|
| | | GPT-4o | Gemini-F | LLama-3.2 | InternVL2.5 | LLama-3.1 | Aya-23 | CLD2 | PP-OCRv3 |
| **Image as the Input** | | | | | | | | | |
| Image Segmentation | IoU ↑ | - | - | - | - | - | - | - | 0.8 |
| OCR | CER ↓ | >1 | >1 | >1 | >1 | - | - | - | >1 |
| Image Transliteration | CER ↓ | >1 | >1 | >1 | >1 | - | - | - | - |
| Image Translation | chrf++ ↑ | 13.0 | 10.0 | 2.9 | 8.6 | - | - | - | - |
| **Local Aksara as the Input** | | | | | | | | | |
| Transliteration | CER ↓ | 0.3 | 0.8 | 1.0 | >1 | >1 | >1 | - | - |
| Translation | chrf++ ↑ | 22.9 | 18.7 | 11.3 | 0.9 | 11.0 | 6.6 | - | - |
| LID | Acc. (%) ↑ | 67.9 | 21.0 | 12.4 | 14.0 | 5.9 | 0.8 | 42.3 | - |
| **Romanized Script as the Input** | | | | | | | | | |
| Translation | chrf++ ↑ | 41.7 | 29.8 | 27.7 | 11.0 | 27.7 | 25.2 | - | - |
| LID | Acc. (%) ↑ | 68.0 | 31.3 | 43.6 | 2.7 | 1.9 | 0.3 | 80.0 | - |

Table 3: Comparative performance of diverse models on multi-modal text tasks (averaged across scripts/languages). The table presents evaluation metrics for various tasks using three input modalities—images, local aksara, and romanized script. Arrows indicate the desired performance direction (↑ higher is better; ↓ lower is better).

**Image Translation** Translating segmented text images directly into Indonesian.

**Language Identification** Identifying languages from both original scripts and their romanized variations. Some sentences consist solely of numbers; therefore, we discard them for LID from romanized scripts.

These task formulations encompass all the language and script data we collect, except for the Lampung language. At the time of writing, Unicode support for Lampung script is unavailable. As a result, no transcription-related tasks are defined for Lampung.

## 4 NusaAksara Benchmark

To evaluate the effectiveness of our dataset and tasks, we conduct a series of experiments using state-of-the-art models across all tasks in NusaAksara benchmark.

### 4.1 Experimental Setup

**Models** As our NusaAksara benchmark covers diverse tasks with both text and image modalities, we employ various models depending on the use cases. Generally, we explore the performance of visual-language models, including both opaque models (GPT-4o (OpenAI et al., 2024), Gemini-Flash (Team et al., 2024)) and publicly available models (Llama-3.2 (Dubey et al., 2024), Intern-VL (Chen et al., 2024), LLaVA-NeXT (Liu et al., 2023)), in a zero-shot manner. We also evaluate multilingual or Indonesian-centric large language models such as Cendol (Cahyawijaya et al., 2024b), BLOOMZ (Muennighoff et al., 2023), Aya (Aryabumi et al., 2024) for task subsets that do not require images as input. We also utilize system-specific models for certain tasks, such as OCR and segmentation (PP-OCR (Du et al., 2020), SAM-ViT(Kirillov et al., 2023)), transliteration (Llama (Dubey et al., 2024)), machine translation (NLLB (Team et al., 2022)), and language identification (CLD2 (Sites, 2013), FastText (Joulin et al., 2017)).

**Metrics** Our metrics also depend on the task. We employ metrics typically used for each specific task. Specifically, we use CER and WER for transliteration and OCR, BLEU (Papineni et al., 2002; Post, 2018) and chrF++ (Popović, 2017) for translation, accuracy for LID, and IoU for image segmentation. However, we only show results with one metric in the main paper due to space constraints, while the rest are included in the Appendix L.

### 4.2 Performance

Table 3 shows the average performance in languages on our NusaAksara benchmark for a selection of models. The results indicate that, in most cases, models struggle with Indonesian local scripts. In contrast, performance is relatively strong when the input is in transliterated text, suggesting that the primary issue lies in the lack of representation of these scripts in the models, as previously discussed in Section 2.1.

**Segmentation and OCR** Both segmentation and OCR performance are shown in Table 4. A fine-tuned PP-OCRv3 based model achieves reasonable segmentation performance. However, DBResNet-50 is lacking, considering that the model was trained on the same dataset and framework. Expectedly, SAM-ViT performs the worst with one-shot experiment setup.

| Model | Sunda | Pegon | Lontara | Jawi | Jawa | Batak | Bali | Lampung |
|---|---|---|---|---|---|---|---|---|
| **Image Segmentation (IoU ↑)** | | | | | | | | |
| PP-OCRv3_det | .59 | .82 | .76 | .89 | .79 | .77 | .91 | .87 |
| SAM-ViT | .05 | .04 | .00 | .04 | .00 | .00 | .04 | .00 |
| DBResNet-50 | .11 | .14 | .09 | .18 | .19 | .38 | .37 | .34 |
| **Transcription from Image – OCR (CER ↓)** | | | | | | | | |
| PP-OCRv3 | >1 | >1 | >1 | >1 | >1 | >1 | >1 | - |
| InternVL2.5-8B | >1 | >1 | >1 | >1 | >1 | >1 | >1 | - |
| LLaVA-V1.6-7B | >1 | >1 | >1 | >1 | >1 | >1 | >1 | - |
| Llama3.2-11B | >1 | >1 | >1 | >1 | >1 | >1 | >1 | - |
| GPT-4o | >1 | >1 | >1 | .44 | >1 | >1 | >1 | - |
| Gemini Flash | >1 | >1 | >1 | >1 | >1 | >1 | >1 | - |

Table 4: Performance on the image segmentation and OCR tasks on various models. For PP-OCRv3 and DBResNet-50 specifically were fine-tuned using PaddleOCR toolkits.

OCR performance is extremely poor. Even when fine-tuned, PP-OCR fails to produce accurate OCR predictions, likely due to the extremely limited training data, which is insufficient for effective learning. All open-source models perform poorly, whereas proprietary models such as GPT-4o and Gemini unexpectedly succeed in OCR for a specific script–Jawi, which is a modified Arabic script used to write the Malay language. However, as shown in Appendix J, these models frequently hallucinate, generating nonsensical text or entirely different scripts, such as Devanagari.

**Transliteration** Open LLMs achieve close to or more than a 100% error rate (i.e., CER of 1) on transliteration in most scripts. Opaque models show significantly better results compared to them, though there is still room for improvement. Again, Jawi is among the scripts where most models perform somewhat well in transliteration. We also see some success with Llama and opaque models on the Jawa script, primarily because it is one of the highest-resource and most widely spoken among Indonesian regional languages. Interestingly, GPT-4o performs decently on the Bali script, while Gemini can't handle it at all.

Transliterating directly from images presents an even greater challenge, as models typically perform worse than when transliterating from the local script. Looking at their outputs, models are hallucinating and producing unrelated texts that are often too long, hence achieving a high CER.

| Model | Sunda | Pegon | Lontara | Jawi | Jawa | Batak | Bali | Lampung |
|---|---|---|---|---|---|---|---|---|
| **Transliteration from Image (CER ↓)** | | | | | | | | |
| InternVL2.5-8B | >1 | >1 | >1 | >1 | >1 | >1 | >1 | >1 |
| LlaVA-v1.6-7B | >1 | >1 | >1 | >1 | >1 | >1 | >1 | >1 |
| Llama3.2-11B | >1 | >1 | >1 | >1 | >1 | >1 | >1 | >1 |
| GPT-4o | >1 | >1 | >1 | .47 | >1 | >1 | >1 | .93 |
| Gemini Flash | >1 | >1 | >1 | .88 | >1 | >1 | .89 | >1 |
| **Transliteration from Local Aksara (CER ↓)** | | | | | | | | |
| Cendol-7b | >1 | >1 | >1 | .86 | >1 | >1 | >1 | - |
| Sailor-7B | >1 | >1 | >1 | .45 | >1 | .69 | >1 | - |
| Bloomz-7B1 | >1 | >1 | >1 | >1 | >1 | >1 | .88 | - |
| Aya-23-8B | >1 | >1 | >1 | .55 | >1 | .91 | >1 | - |
| Llama-3.1-8B | >1 | >1 | .66 | .42 | >1 | .97 | .89 | - |
| Lama-3.2-11B | .77 | .87 | > 1 | .41 | 0.61 | >1 | 1.0 | - |
| InternVL2.5-8B | >1 | >1 | >1 | >1 | >1 | >1 | >1 | - |
| GPT-4o | .17 | .33 | .31 | .2 | .28 | .82 | .33 | - |
| Gemini Flash | .58 | >1 | .64 | .31 | .32 | .9 | >1 | - |

Table 5: Character Error Rate (CER) comparison across models for image-based and aksara-based transliteration (the lower, the better).

| Model | ban | btx | jav$_{jj}$ | zsm | bug | jav$_{jp}$ | sun |
|---|---|---|---|---|---|---|---|
| **Translation from Image (ChrF++ ↑)** | | | | | | | |
| GPT-4o | 11.2 | 8.9 | 12.4 | 30.9 | 10.5 | 12.6 | 11.0 |
| Gemini Flash | 15.7 | 4.6 | 11.0 | 17.5 | 9.7 | 7.4 | 9.8 |
| InternVL2.5-8B | 14.1 | 4.5 | 9.1 | 12.0 | 8.9 | 7.5 | 7.8 |
| LLaVA-v1.6-7B | 8.3 | 1.3 | 5.3 | 4.3 | 4.7 | 3.9 | 3.7 |
| Llama3.2-11B | 4.8 | 1.2 | 2.9 | 4.4 | 3.1 | 2.6 | 2.9 |
| **Translation from Local Aksara(ChrF++ ↑)** | | | | | | | |
| Cendol | 11.6 | 5.3 | 11.3 | 13.2 | 12.3 | 9.6 | 11.3 |
| Sailor-7B | 7.0 | 2.2 | 6.3 | 12.0 | 5.0 | 4.2 | 4.8 |
| bloomz-7b1 | 11.1 | 10.1 | 12.3 | 12.3 | 13.4 | 7.2 | 11.4 |
| aya-23-8B | 4.8 | 4.0 | 5.5 | 13.9 | 7.5 | 4.0 | 6.6 |
| Llama-3.1-8B | 12.4 | 7.5 | 9.7 | 19.7 | 13.3 | 5.2 | 9.5 |
| Llama-3.2-11B | 12.7 | 8.6 | 9.8 | 19.9 | 13.2 | 5.3 | 9.7 |
| GPT-4o | 15.6 | 7.7 | 18.0 | 48.9 | 12.9 | 24.3 | 20.5 |
| Gemini | 12.4 | 5.9 | 16.2 | 42.3 | 13.3 | 21.3 | 13.2 |
| NLLB-3.3B | 2.8 | 2.3 | 3.6 | 20.8 | 9.3 | 5.2 | 6.9 |
| InternVL2.5-8B | 0.1 | 0.0 | 0.7 | 2.2 | 0.4 | 1.4 | 1.4 |
| **Translation from Romanized Script (ChrF++ ↑)** | | | | | | | |
| Cendol | 19.1 | 27.9 | 35.6 | 43.4 | 16.8 | 28.8 | 34.5 |
| Sailor-7B | 14.0 | 32.1 | 23.5 | 41.9 | 16.1 | 20.1 | 23.8 |
| bloomz-7b1 | 13.8 | 22.4 | 18.2 | 39.8 | 14.0 | 16.4 | 19.1 |
| aya-23-8B | 14.0 | 29.7 | 23.1 | 42.5 | 14.9 | 19.5 | 23.9 |
| Llama-3.1-8B | 15.5 | 28.6 | 23.1 | 39.6 | 16.2 | 26.1 | 25.5 |
| Llama-3.2-11B | 15.5 | 28.4 | 23.1 | 38.7 | 16.4 | 26.4 | 25.3 |
| GPT-4o | 27.5 | 34.2 | 46.3 | 58.5 | 22.8 | 50.2 | 48.0 |
| Gemini | 23.6 | 23.8 | 37.4 | 49.0 | 16.8 | 19.8 | 37.5 |
| NLLB-3.3B | 20.2 | 32.0 | 33.7 | 48.9 | 24.1 | 31.4 | 36.5 |
| InternVL2.5-8B | 11.4 | 7.5 | 12.0 | 18.9 | 10.5 | 8.8 | 11.1 |

Table 6: ChrF++ performance of various models on different languages for translation tasks.

**Translation** As expected, translating from romanized script is decent in some languages. In contrast, translating directly from the local script is challenging. Similar to transliteration, only opaque models have some capability in this regard. Their performance on the Jawi script is notably higher; however, it remains subpar.

**LID** Language identification (LID) is one of the few tasks where models do not perform as poorly. Some popular LID toolkits can accurately identify languages, even when presented with local scripts. We argue that this task may be easier because most scripts are uniquely associated with specific languages. However, an exception is the Jawi and Pegon scripts, which are used for Malay and Javanese but share similarities with Arabic. The low performance in this case is due to LID models misclassifying text written in Jawi or Pegon as Arabic. LID performance deteriorates further for romanized scripts, as models are undertrained for these languages, resulting in poor accuracy. Notably, GPT-4o is performing well, whereas Gemini is almost always predicting Javanese.

| Model | ban | btx | $jav_{jj}$ | zsm | bug | $jav_{jp}$ | sun |
|---|---|---|---|---|---|---|---|
| **LID on Romanized Script (%)** | | | | | | | |
| LangID | 0 | - | 40.7 | 0 | - | 0 | - |
| fasttext | - | - | 34.5 | - | - | 0 | 18.3 |
| CLD2 | - | - | 42.0 | - | - | 0 | 42.6 |
| GPT-4o | 99.4 | 100 | 99.8 | 42.32 | 34.31 | 0 | 100 |
| Gemini | 0.4 | 0.9 | 99.5 | 13.1 | 5.2 | 0 | 100 |
| **LID on Local Aksara (%)** | | | | | | | |
| LangID | 0 | - | 0 | 0 | - | 0 | - |
| fasttext | - | - | 0 | 0 | - | 0 | 0 |
| CLD2 | 86.5 | 100 | 98.7 | - | 95.4 | 0 | 98.8 |
| GPT-4o | 99.4 | 100 | 99.8 | 42.3 | 34.3 | 0 | 100 |
| Gemini | 9.2 | 6.7 | 84.0 | 0.1 | 0 | 0 | 47.43 |

Table 7: Language Identification accuracy

## 5 Related Work

**Preserving Low-Resource and Endangered Languages** Language preservation efforts have mainly targeted marginalized spoken languages (Bird, 2020; McMillan-Major et al., 2022; Zhang et al., 2023). While large multilingual initiatives like XTREME-R use cross-lingual transfer to accelerate development (Hu et al., 2020; Clark et al., 2020; Liang et al., 2020; Ruder et al., 2021), they typically focus on languages with robust digital support, leaving traditional scripts largely neglected (Littell et al., 2018; Zhong et al., 2024).

**Multilingual and Regional Language Benchmarks** Multilingual benchmarks such as XNLI, MLQA, TyDiQA, and XGLUE cover a wide range of languages (Conneau et al., 2018; Lewis et al., 2020; Clark et al., 2020; Liang et al., 2020), while regional collections such as MasakhaNER, AmericasNLI, and Samanantar enhance representation (Adelani et al., 2022; Ebrahimi et al., 2022; Ramesh et al., 2023). Similarly, efforts in South Asia such as IndicNLP, IndicCorp and Southeast Asia such as IndoNLU, NusaWrites, NusaX have strengthened local language resources (Kakwani et al., 2020; Kunchukuttan et al., 2020; Wilie et al., 2020; Cahyawijaya et al., 2023; Winata et al., 2023). Arabic-script varieties also benefit from ARBENCH (Abdul-Mageed et al., 2021). While efforts have been made to create benchmarks for Indonesian languages, they often rely on romanized scripts, neglecting endangered writing systems and historical orthographies (Schwenk et al., 2021; Agić and Vulić, 2019; El-Kishky et al., 2020).

**Digital Infrastructure for Scripts** Digitizing historical scripts remains a challenge, especially in Southeast Asia, where complex characters and limited Unicode support hinder preservation (Areni et al., 2017; Mudiarta et al., 2020). Projects like DREAMSEA (Dreamsea, 2024), the Southeast Asia Digital Library (Berkeley, 2023), Nusantara Scripts OCR (Prasetiadi et al., 2023), and Hán Nôm digitization (Van Phan et al., 2015) have made strides. Tools such as Aksharamukha (Rajan, 2024) help in script conversion, yet there are gaps and incomplete standards that require culturally informed digitization (Purwarianti et al., 2025).

## 6 Conclusion

We constructed a novel dataset, NusaAksara, for Indonesian languages that focuses on indigenous scripts across multiple tasks, including image segmentation, OCR, transliteration, translation, and Language Identification (LID). Curated from local manuscripts and carefully annotated and validated by experts, NusaAksara brings attention to the huge gap in existing NLP resources, which are still heavily relied toward romanized text. By evaluating various models on NusaAksara, we found that most NLP systems struggle with these non-Latin scripts, thus represent the urgent need for broader support. Our findings reveal the urgent need of integrating indigenous scripts into NLP pipelines to encourage linguistic preservation and improved accessibility for historically marginalized scripts and languages.

## Limitations

This study observed only eight of the 20 recognized local scripts, and the lack of Unicode support for Lampung scripts presents a significant challenge for transcription-related pipelines such as OCR, transliteration, and translation of local scripts. Although efforts have been made to incorporate Lampung scripts into Unicode, they have not yet been officially supported at the time of writing. Additionally, due to book content copyrights and in compliance with ethical guidelines, we were only able to annotate and provide 10% of the available resources; gathering more resources would be beneficial for the further development of NusaAksara.

## A Script of Focus

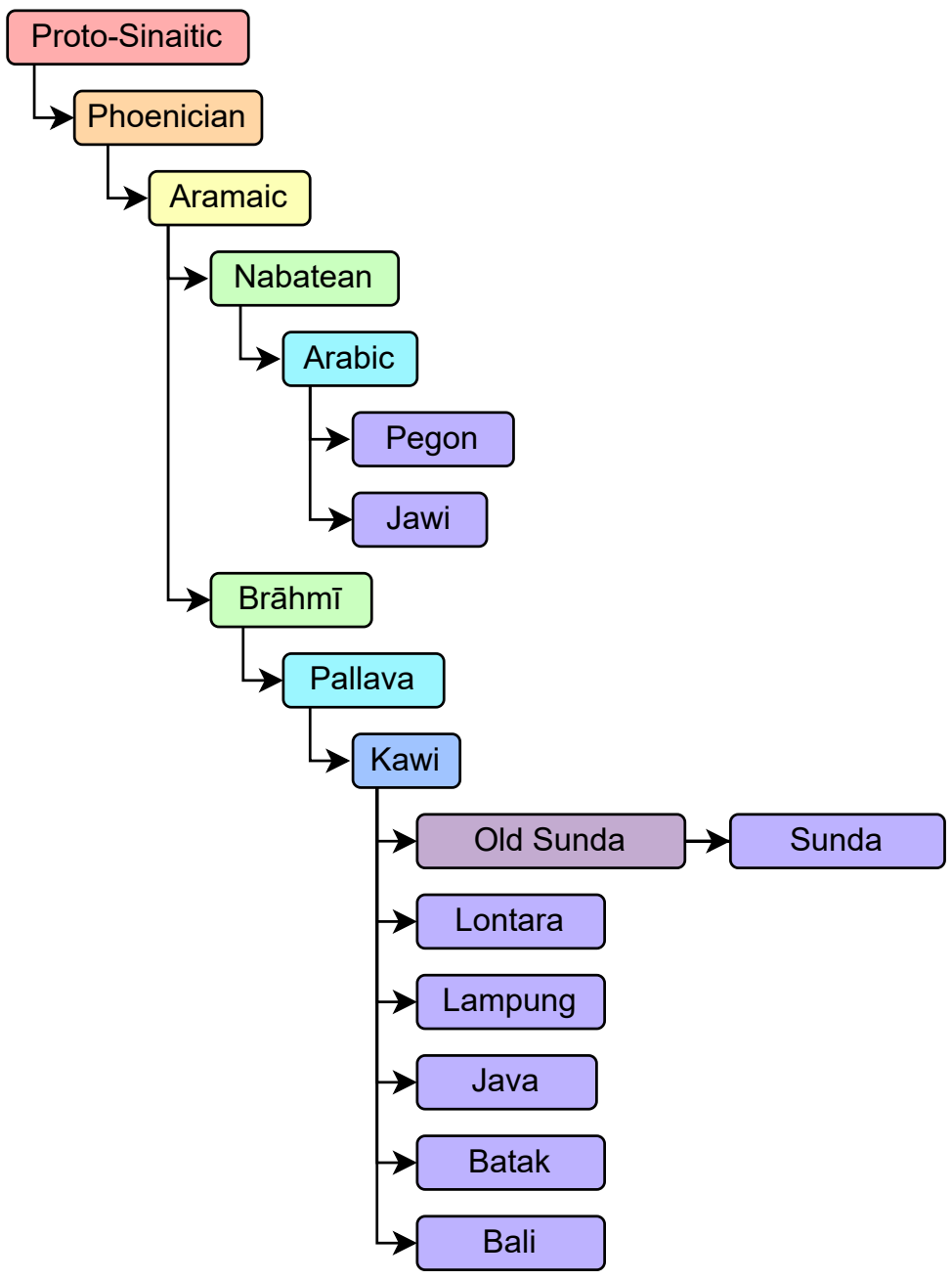

Figure 3: The script taxonomy for the eight focus local aksara based on Omniglot (Ager, 2002). In this taxonomy, the color indicates the category level of the language, with purple representing the specific language and various other colors correspond to the language family.

Below, we provide an overview of the languages, their scripts, approximate number of speakers,[7] and key linguistic features.

**Aksara Bali (ban).** Balinese is an Austronesian language spoken primarily on the island of Bali and in parts of West Nusa Tenggara. It has around 3–3.5 million speakers. While most modern Balinese texts are written in the Latin script, the traditional Bali script—derived from the Brahmi family—is still taught and used for ceremonial or literary purposes. Balinese has three sociolinguistic registers (often called *levels of speech*), reflecting differences in formality and the social status of the interlocutor (CLYNES, 2007). Its basic word order is SVO, and it has a rich system of affixation, including prefixes, suffixes, circumfixes, and reduplication.

**Aksara Batak (btx, bbc).** Aksara Batak is commonly used across several Batak langauges, among them are:

- *Batak Karo* (btx), spoken by approximately 600,000–700,000 people in North Sumatras Karo highlands.
- *Batak Toba* (bbc), with around 2 million speakers primarily around Lake Toba in North Sumatra.

Both traditionally use the **Batak script** (Surat Batak), a Brahmic-derived script. Modern usage predominantly relies on the Latin alphabet. Batak languages are often described as having verb-initial structures with rich verbal morphology reminiscent of Philippine-type languages, though they differ in many details (Blust, 2013). They have also been influenced by neighboring Malayic languages and Indonesian due to commerce and migration.

**Aksara Jawa (jav).** Javanese is the largest Austronesian language in Indonesia by number of native speakers, estimated at 82–85 million (Eberhard et al., 1997). Its traditional script, Aksara Jawa, is a Brahmic-derived script still taught in schools in Central and East Java, though its practical use is limited compared to Latin script. Javanese has at least three major speech levels: *Ngoko*, *Krama*, and *Krama*

[7] Speaker estimates are derived from Ethnologue (Eberhard et al., 1997) and various regional sources.

*Inggil*, which reflect social hierarchy and formality (Isodarus, 2020; Wedhawati et al., 2001). The language employs a basic SVO word order, but with extensive voice and affixation systems.

**Aksara Jawi (zsm).** **Jawi** is the Arabic-derived script used primarily for Malay (zsm), but also for writing Arabic (arb) texts in the Southeast Asian context. Historically, Jawi was used throughout the Malay-speaking world (including parts of Sumatra, the Malay Peninsula, and coastal Borneo). Contemporary usage is more common in religious or traditional contexts. Modern Malay and Indonesian both share a high degree of mutual intelligibility, and Jawi sees continued but limited use in certain regions (e.g., Brunei, parts of Malaysia, and Indonesian pesantren).

**Aksara Lampung (ljp).** Lampung is an Austronesian language native to the Lampung province in southern Sumatra, spoken by around 1.5 million people. It traditionally employs the **Lampung script** (Aksara Lampung), another Brahmic-based abugida also known as *Ka Ga Nga*. Currently, many speakers predominantly use the Latin script, and language shift towards Indonesian is common. Lampung has several dialects (e.g., Nyow and Abung) and exhibits typical Austronesian features such as affixation and reduplication, with an SVO word order.

**Aksara Lontara (bug).** **Buginese** (bug) is the language of around 5 million speakers in South Sulawesi. The traditional **Lontara** script is a Brahmic-derived abugida closely related to other South Sulawesi scripts. Although it remains a cultural symbol, modern Buginese writing is more often in the Latin script. Buginese has a rich morphology, including person-marking on verbs, and typically follows SVO word order. Politeness or deference in speech is conveyed through choice of pronouns, affixes, and lexicon (Weda, 2016).

**Aksara Pegon (jav).** **Pegon** is the adaptation of the Arabic script for writing the Javanese language, though it can also be used for Arabic quotes or terms embedded in Javanese texts. Similar to Jawi for Malay, Pegon has been historically significant in Islamic boarding schools across Java for religious and educational texts. Despite being overshadowed by Latin-based Javanese today, Pegon still holds cultural importance in traditional religious literature and local Islamic contexts.

**Aksara Sunda (sun).** Sundanese is an Austronesian language spoken by around 39 million people in West Java and Banten. Its classical form used the **Sundanese script** (Aksara Sunda), another Brahmic-based writing system, though Latin script prevails in modern times. Sundanese exhibits SVO word order, a voice-marking system similar to that in Indonesian, and elaborate registers for conveying respect (Kurniawan, 2013). Historically, it was also written in *Pegon* (modified Arabic script) for religious texts, underscoring its capacity for diverse orthographic representations.

## B Script Distribution

| Model Names | Script Names | Number of Unique Tokens | Percentage of Unique Tokens (%) | Number of Tokens | Percentage of Tokens (%) |
|---|---|---|---|---|---|
| facebook/nllb-200-3,3B | Latin | 138,482 | 54.0519 | 17,416,796,244 | 53.0683 |
| | Cyrillic | 22,686 | 8.8547 | 2,815,751,150 | 8.5795 |
| | Arabic | 13,997 | 5.4633 | 1,677,366,993 | 5.1109 |
| | Japanese | 11,228 | 4.3825 | 1,730,543,337 | 5.2729 |
| | Devanagari | 8,404 | 3.2802 | 984,860,186 | 3.0008 |
| | Hangul | 7,985 | 3.1167 | 1,150,210,268 | 3.5046 |
| | Non-Language Specific | 5,650 | 2.2053 | 771,508,749 | 2.3508 |
| | Bengali | 3,938 | 1.5371 | 489,980,917 | 1.493 |
| | Ethiopic | 3,632 | 1.4176 | 508,035,159 | 1.548 |
| | Greek | 3,109 | 1.2135 | 390,617,504 | 1.1902 |
| | Hebrew | 3,090 | 1.2061 | 385,535,367 | 1.1747 |
| | Gujarati | 2,614 | 1.0203 | 332,137,051 | 1.012 |
| | Telugu | 2,511 | 0.9801 | 316,251,033 | 0.9636 |
| | Tibetan | 2,494 | 0.9735 | 301,026,275 | 0.9172 |
| | Kannada | 2,480 | 0.968 | 311,335,963 | 0.9486 |
| | Malayalam | 2,378 | 0.9282 | 298,607,617 | 0.9098 |
| | Oriya | 2,223 | 0.8677 | 273,639,606 | 0.8338 |
| | Tamil | 2,196 | 0.8571 | 274,202,982 | 0.8355 |
| | Armenian | 2,130 | 0.8314 | 269,067,058 | 0.8198 |
| | Myanmar | 1,979 | 0.7724 | 245,776,967 | 0.7489 |
| | Georgian | 1,962 | 0.7658 | 252,388,118 | 0.769 |
| | Gurmukhi | 1,829 | 0.7139 | 229,288,070 | 0.6986 |
| | Thai | 1,665 | 0.6499 | 206,573,997 | 0.6294 |
| | Sinhala | 1,616 | 0.6308 | 201,175,458 | 0.613 |
| | Lao | 1,539 | 0.6007 | 192,654,149 | 0.587 |
| | Khmer | 1,513 | 0.5905 | 190,593,959 | 0.5807 |
| | Traditional Chinese | 1,373 | 0.5359 | 294,353,930 | 0.8969 |
| | Simplified Chinese | 1,030 | 0.402 | 233,917,322 | 0.7127 |
| | Tifinagh Script | 259 | 0.1011 | 39,133,299 | 0.1192 |
| | Ol Chiki Script | 172 | 0.0671 | 27,292,451 | 0.0832 |
| | Unknown Script | 38 | 0.0148 | 8,991,607 | 0.0274 |
| bigscience/bloomz-7b1 | Latin | 119,450 | 47.7115 | 14,756,213,993 | 47.0107 |
| | Japanese | 25,758 | 10.2884 | 3,480,599,313 | 11.0886 |
| | Arabic | 20,590 | 8.2242 | 2,640,386,762 | 8.4118 |
| | Devanagari | 15,920 | 6.3589 | 1,969,385,166 | 6.2741 |
| | Non-Language Specific | 10,917 | 4.3605 | 1,247,277,162 | 3.9736 |
| | Bengali | 10,562 | 4.2187 | 1,340,439,559 | 4.2704 |
| | Telugu | 6,462 | 2.5811 | 835,932,657 | 2.6631 |
| | Kannada | 6,361 | 2.5408 | 824,452,581 | 2.6266 |

| Model Names | Script Names | Number of Unique Tokens | Percentage of Unique Tokens (%) | Number of Tokens | Percentage of Tokens (%) |
|---|---|---|---|---|---|
| | Tamil | 6,195 | 2.4744 | 784,360,210 | 2.4988 |
| | Malayalam | 5,891 | 2.353 | 771,506,477 | 2.4579 |
| | Gujarati | 5,627 | 2.2476 | 716,698,853 | 2.2833 |
| | Gurmukhi | 5,274 | 2.1066 | 668,586,735 | 2.13 |
| | Oriya | 4,722 | 1.8861 | 602,045,062 | 1.918 |
| | Simplified Chinese | 2,838 | 1.1336 | 293,064,744 | 0.9337 |
| | Traditional Chinese | 2,191 | 0.8751 | 237,652,774 | 0.7571 |
| | Cyrillic | 727 | 0.2904 | 96,157,735 | 0.3063 |
| | Hangul | 342 | 0.1366 | 57,726,299 | 0.1839 |
| | Greek | 195 | 0.0779 | 23,543,716 | 0.075 |
| | Unknown Script | 117 | 0.0467 | 10,837,106 | 0.0345 |
| | Armenian | 56 | 0.0224 | 7,346,477 | 0.0234 |
| | Hebrew | 53 | 0.0212 | 7,509,611 | 0.0239 |
| | Thai | 42 | 0.0168 | 5,804,436 | 0.0185 |
| | Georgian | 24 | 0.0096 | 3,295,705 | 0.0105 |
| | Khmer | 14 | 0.0056 | 2,539,842 | 0.0081 |
| | Coptic | 12 | 0.0048 | 2,369,817 | 0.0075 |
| | Yi | 6 | 0.0024 | 915,770 | 0.0029 |
| | Gothic | 5 | 0.002 | 799,851 | 0.0025 |
| | Tibetan | 3 | 0.0012 | 610,252 | 0.0019 |
| | Mongolian | 3 | 0.0012 | 559,571 | 0.0018 |
| | Ethiopic | 1 | 0.0004 | 245,407 | 0.0008 |
| | Undefined Chinese | 1 | 0.0004 | 222,408 | 0.0007 |
| indonlp/cendol-mt5-large-inst | Latin | 116,712 | 46.6665 | 13,294,675,679 | 42.5092 |
| | Cyrillic | 26,685 | 10.6698 | 3,166,559,640 | 10.125 |
| | Non-Language Specific | 22,127 | 8.8473 | 3,250,482,912 | 10.3933 |
| | Japanese | 21,733 | 8.6898 | 3,548,133,754 | 11.345 |
| | Arabic | 7,226 | 2.8893 | 615,516,308 | 1.9681 |
| | Greek | 5,217 | 2.086 | 590,104,485 | 1.8868 |
| | Thai | 4,391 | 1.7557 | 664,809,908 | 2.1257 |
| | Hangul | 4,126 | 1.6498 | 518,299,050 | 1.6572 |
| | Hebrew | 4,036 | 1.6138 | 384,282,950 | 1.2287 |
| | Tamil | 3,298 | 1.3187 | 453,041,660 | 1.4486 |
| | Devanagari | 3,075 | 1.2295 | 294,002,442 | 0.9401 |
| | Malayalam | 2,948 | 1.1787 | 428,519,064 | 1.3702 |
| | Simplified Chinese | 2,783 | 1.1128 | 466,061,105 | 1.4902 |
| | Georgian | 2,589 | 1.0352 | 331,992,752 | 1.0615 |
| | Traditional Chinese | 2,547 | 1.0184 | 495,604,879 | 1.5847 |

| Model Names | Script Names | Number of Unique Tokens | Percentage of Unique Tokens (%) | Number of Tokens | Percentage of Tokens (%) |
|---|---|---|---|---|---|
| | Telugu | 2,346 | 0.938 | 310,769,318 | 0.9937 |
| | Myanmar | 2,279 | 0.9112 | 349,412,913 | 1.1172 |
| | Armenian | 2,261 | 0.904 | 270,515,705 | 0.865 |
| | Kannada | 2,155 | 0.8617 | 286,393,459 | 0.9157 |
| | Khmer | 1,976 | 0.7901 | 312,657,642 | 0.9997 |
| | Bengali | 1,787 | 0.7145 | 165,327,379 | 0.5286 |
| | Sinhala | 1,679 | 0.6713 | 162,399,247 | 0.5193 |
| | Lao | 1,412 | 0.5646 | 220,628,291 | 0.7055 |
| | Unknown Script | 1,361 | 0.5442 | 324,644,774 | 1.038 |
| | Gujarati | 1,108 | 0.443 | 105,652,303 | 0.3378 |
| | Ethiopic | 1,004 | 0.4014 | 91,452,942 | 0.2924 |
| | Gurmukhi | 571 | 0.2283 | 34,015,793 | 0.1088 |
| | Canadian Aboriginal Syllabics | 89 | 0.0356 | 21,900,957 | 0.07 |
| | Thaana | 83 | 0.0332 | 14,089,740 | 0.0451 |
| | Oriya | 83 | 0.0332 | 9,148,262 | 0.0293 |
| | Unmapped Script | 46 | 0.0184 | 11,335,317 | 0.0362 |
| | Mongolian | 45 | 0.018 | 6,941,581 | 0.0222 |
| | Tibetan | 39 | 0.0156 | 8,509,958 | 0.0272 |
| | Tifinagh Script | 32 | 0.0128 | 7,446,718 | 0.0238 |
| | Syriac | 32 | 0.0128 | 6,678,874 | 0.0214 |
| | Coptic | 30 | 0.012 | 7,201,011 | 0.023 |
| | Balinese | 26 | 0.0104 | 6,314,994 | 0.0202 |
| | Runic Script | 26 | 0.0104 | 6,403,422 | 0.0205 |
| | Cherokee Script | 25 | 0.01 | 6,195,045 | 0.0198 |
| | Shavian | 18 | 0.0072 | 4,404,745 | 0.0141 |
| | Newa | 18 | 0.0072 | 4,438,134 | 0.0142 |
| | N'Ko | 14 | 0.0056 | 3,214,595 | 0.0103 |
| | Cham | 11 | 0.0044 | 2,535,124 | 0.0081 |
| | Rejang | 6 | 0.0024 | 1,469,639 | 0.0047 |
| | Gothic | 6 | 0.0024 | 1,489,129 | 0.0048 |
| | Yi | 6 | 0.0024 | 1,483,034 | 0.0047 |
| | Tai Scripts | 5 | 0.002 | 1,219,633 | 0.0039 |
| | Buginese | 4 | 0.0016 | 982,641 | 0.0031 |
| | Brahmi Script | 4 | 0.0016 | 997,329 | 0.0032 |
| | Mandaic Script | 4 | 0.0016 | 986,865 | 0.0032 |
| | Ol Chiki Script | 3 | 0.0012 | 739,375 | 0.0024 |
| | Samaritan Script | 3 | 0.0012 | 743,832 | 0.0024 |
| | Undefined Chinese | 3 | 0.0012 | 737,143 | 0.0024 |

| Model Names | Script Names | Number of Unique Tokens | Percentage of Unique Tokens (%) | Number of Tokens | Percentage of Tokens (%) |
|---|---|---|---|---|---|
| | Kayah Li Script | 2 | 0.0008 | 487,708 | 0.0016 |
| | Lisu | 1 | 0.0004 | 249,943 | 0.0008 |
| | Ogham Script | 1 | 0.0004 | 248,305 | 0.0008 |
| | Sundanese | 1 | 0.0004 | 249,822 | 0.0008 |
| meta-llama/Llama-3,1-8B-Instruct | Latin | 97,272 | 76.1568 | 5,403,516,373 | 65.921 |
| | Non-Language Specific | 8,801 | 6.8905 | 449,387,485 | 5.4824 |
| | Cyrillic | 6,515 | 5.1008 | 702,459,906 | 8.5698 |
| | Japanese | 4,070 | 3.1865 | 427,293,684 | 5.2128 |
| | Arabic | 3,714 | 2.9078 | 416,823,558 | 5.0851 |
| | Hangul | 2,289 | 1.7921 | 248,007,013 | 3.0256 |
| | Greek | 1,392 | 1.0898 | 155,970,486 | 1.9028 |
| | Thai | 1,346 | 1.0538 | 149,911,828 | 1.8289 |
| | Devanagari | 905 | 0.7085 | 100,194,470 | 1.2223 |
| | Simplified Chinese | 812 | 0.6357 | 79,339,769 | 0.9679 |
| | Traditional Chinese | 495 | 0.3875 | 55,762,720 | 0.6803 |
| | Unknown Script | 89 | 0.0697 | 6,428,477 | 0.0784 |
| | Hebrew | 22 | 0.0172 | 1,459,279 | 0.0178 |
| | Armenian | 2 | 0.0016 | 237,192 | 0.0029 |
| | Bengali | 2 | 0.0016 | 161,006 | 0.002 |
| meta-llama/Llama-3,2-11B-Vision-Instruct | Latin | 97,273 | 76.157 | 5,403,644,629 | 65.9216 |
| | Non-Language Specific | 8,801 | 6.8905 | 449,387,485 | 5.4823 |
| | Cyrillic | 6,515 | 5.1007 | 702,459,906 | 8.5696 |
| | Japanese | 4,070 | 3.1865 | 427,293,684 | 5.2128 |
| | Arabic | 3,714 | 2.9078 | 416,823,558 | 5.085 |
| | Hangul | 2,289 | 1.7921 | 248,007,013 | 3.0256 |
| | Greek | 1,392 | 1.0898 | 155,970,486 | 1.9028 |
| | Thai | 1,346 | 1.0538 | 149,911,828 | 1.8288 |
| | Devanagari | 905 | 0.7085 | 100,194,470 | 1.2223 |
| | Simplified Chinese | 812 | 0.6357 | 79,339,769 | 0.9679 |
| | Traditional Chinese | 495 | 0.3875 | 55,762,720 | 0.6803 |
| | Unknown Script | 89 | 0.0697 | 6,428,477 | 0.0784 |
| | Hebrew | 22 | 0.0172 | 1,459,279 | 0.0178 |
| | Bengali | 2 | 0.0016 | 161,006 | 0.002 |
| | Armenian | 2 | 0.0016 | 237,192 | 0.0029 |
| sail/Sailor-7B | Latin | 94,601 | 62.5647 | 5,117,161,765 | 44.5718 |
| | Japanese | 22,203 | 14.684 | 2,476,541,565 | 21.5713 |
| | Non-Language Specific | 10,332 | 6.8331 | 836,140,509 | 7.283 |
| | Simplified Chinese | 4,281 | 2.8313 | 468,385,962 | 4.0798 |

| Model Names | Script Names | Number of Unique Tokens | Percentage of Unique Tokens (%) | Number of Tokens | Percentage of Tokens (%) |
|---|---|---|---|---|---|
| | Cyrillic | 4,149 | 2.744 | 502,418,700 | 4.3762 |
| | Arabic | 3,979 | 2.6315 | 530,145,901 | 4.6177 |
| | Hangul | 3,585 | 2.371 | 486,885,190 | 4.2409 |
| | Hebrew | 3,183 | 2.1051 | 422,949,371 | 3.684 |
| | Thai | 2,540 | 1.6798 | 334,303,820 | 2.9119 |
| | Traditional Chinese | 921 | 0.6091 | 104,013,125 | 0.906 |
| | Greek | 232 | 0.1534 | 30,300,701 | 0.2639 |
| | Undefined Chinese | 202 | 0.1336 | 29,481,565 | 0.2568 |
| | Ethiopic | 112 | 0.0741 | 16,752,751 | 0.1459 |
| | Armenian | 73 | 0.0483 | 10,765,618 | 0.0938 |
| | Canadian Aboriginal Syllabics | 71 | 0.047 | 10,618,019 | 0.0925 |
| | Devanagari | 56 | 0.037 | 7,187,368 | 0.0626 |
| | Tai Scripts | 43 | 0.0284 | 6,457,193 | 0.0562 |
| | Unknown Script | 42 | 0.0278 | 1,082,132 | 0.0094 |
| | Bengali | 39 | 0.0258 | 5,645,461 | 0.0492 |
| | Georgian | 36 | 0.0238 | 5,310,217 | 0.0463 |
| | Myanmar | 36 | 0.0238 | 5,358,292 | 0.0467 |
| | Khmer | 33 | 0.0218 | 4,882,543 | 0.0425 |
| | Lao | 33 | 0.0218 | 4,878,698 | 0.0425 |
| | N'Ko | 32 | 0.0212 | 4,754,037 | 0.0414 |
| | Malayalam | 31 | 0.0205 | 4,615,950 | 0.0402 |
| | Mongolian | 28 | 0.0185 | 4,196,956 | 0.0366 |
| | Coptic | 27 | 0.0179 | 4,026,670 | 0.0351 |
| | Syriac | 26 | 0.0172 | 3,830,798 | 0.0334 |
| | Kannada | 25 | 0.0165 | 3,737,893 | 0.0326 |
| | Sinhala | 25 | 0.0165 | 3,720,035 | 0.0324 |
| | Tamil | 25 | 0.0165 | 3,697,418 | 0.0322 |
| | Tibetan | 25 | 0.0165 | 3,711,822 | 0.0323 |
| | Tifinagh Script | 25 | 0.0165 | 3,700,032 | 0.0322 |
| | Javanese | 18 | 0.0119 | 2,689,019 | 0.0234 |
| | Gujarati | 16 | 0.0106 | 2,391,319 | 0.0208 |
| | Cherokee Script | 15 | 0.0099 | 2,243,047 | 0.0195 |
| | Telugu | 14 | 0.0093 | 2,092,466 | 0.0182 |
| | Runic Script | 12 | 0.0079 | 1,796,962 | 0.0157 |
| | Gothic | 10 | 0.0066 | 1,508,593 | 0.0131 |
| | Gurmukhi | 10 | 0.0066 | 1,493,489 | 0.013 |
| | Yi | 10 | 0.0066 | 1,494,915 | 0.013 |
| | Thaana | 8 | 0.0053 | 1,198,039 | 0.0104 |

| Model Names | Script Names | Number of Unique Tokens | Percentage of Unique Tokens (%) | Number of Tokens | Percentage of Tokens (%) |
|---|---|---|---|---|---|
| | Oriya | 7 | 0.0046 | 1,051,622 | 0.0092 |
| | Mandaic Script | 6 | 0.004 | 883,718 | 0.0077 |
| | Buginese | 5 | 0.0033 | 750,360 | 0.0065 |
| | Bamum Script | 4 | 0.0026 | 603,004 | 0.0053 |
| | Limbu Script | 3 | 0.002 | 451,744 | 0.0039 |
| | Samaritan Script | 3 | 0.002 | 452,402 | 0.0039 |
| | Ogham Script | 3 | 0.002 | 450,096 | 0.0039 |
| | Balinese | 2 | 0.0013 | 300,789 | 0.0026 |
| | Modi Script | 1 | 0.0007 | 151,267 | 0.0013 |
| | Sundanese | 1 | 0.0007 | 149,590 | 0.0013 |
| | Lepcha Script | 1 | 0.0007 | 149,594 | 0.0013 |
| | Lisu | 1 | 0.0007 | 150,825 | 0.0013 |
| | Kaithi Script | 1 | 0.0007 | 151,265 | 0.0013 |
| | Ol Chiki Script | 1 | 0.0007 | 150,580 | 0.0013 |
| | Batak Script | 1 | 0.0007 | 149,592 | 0.0013 |
| | Vai Script | 1 | 0.0007 | 148,775 | 0.0013 |
| CohereForAI/aya-23-8B | Latin | 174,122 | 68.4047 | 21,956,668,778 | 67.621 |
| | Cyrillic | 25,060 | 9.8449 | 3,360,867,624 | 10.3506 |
| | Japanese | 19,204 | 7.5444 | 2,698,788,307 | 8.3116 |
| | Greek | 7,557 | 2.9688 | 1,023,756,897 | 3.1529 |
| | Hangul | 6,866 | 2.6973 | 954,231,410 | 2.9388 |
| | Arabic | 6,590 | 2.5889 | 891,352,513 | 2.7451 |
| | Non-Language Specific | 6,253 | 2.4565 | 479,648,107 | 1.4772 |
| | Hebrew | 4,194 | 1.6476 | 581,572,678 | 1.7911 |
| | Simplified Chinese | 1,991 | 0.7822 | 218,554,328 | 0.6731 |
| | Traditional Chinese | 1,705 | 0.6698 | 197,100,391 | 0.607 |
| | Devanagari | 820 | 0.3221 | 91,852,600 | 0.2829 |
| | Unknown Script | 95 | 0.0373 | 2,028,277 | 0.0062 |
| | Thai | 39 | 0.0153 | 5,497,192 | 0.0169 |
| | Armenian | 15 | 0.0059 | 2,084,580 | 0.0064 |
| | Georgian | 13 | 0.0051 | 1,756,307 | 0.0054 |
| | Tamil | 10 | 0.0039 | 1,899,428 | 0.0058 |
| | Bengali | 9 | 0.0035 | 1,670,143 | 0.0051 |
| | Myanmar | 2 | 0.0008 | 467,744 | 0.0014 |
| | Khmer | 1 | 0.0004 | 194,031 | 0.0006 |
| | Tibetan | 1 | 0.0004 | 219,129 | 0.0007 |

## C Prompts of Tasks

The following are the prompts that we used for our experiment.

| Task Name | Task Prompt |
|---|---|
| Script Identification | Answer with only the language name.<br><br>What script is this text written in? |
| Language Identification | Answer with only the language name.<br><br>What language is this text written in? |
| Image Transcription | Answer only with the transcription.<br><br>Transcript this image of [LANG] text script: |
| Image Translation | Only answer with the Indonesian translation.<br><br>Translate this image of [LANG] text script into Indonesian: |
| Image Transliteration | Answer only with the transliteration.<br><br>Transliterate this image of [LANG] text script: |
| Transcription Translation (Aksara to Indo) | Answer only with the translated text.<br><br>Translate this text from its script to Indonesian: [TRANSCRIPTION] |
| Transliteration (Aksara to Latin) | Answer only with the transliteration.<br><br>Convert this script text into Latin: [TRANSCRIPTION] |
| Transliteration Translation (Latin to Indo) | Answer only with the translated text.<br><br>Translate this Latin-transliterated text into Indonesian: [TRANSLITERATION] |

Table 8: Task prompts for different language processing tasks.

## D Downstream Task Script Coverage

In SEACrowd, one of the biggest data catalogue for Southeast Asia, including Indonesian languages, only 2 of them are written in the local script.

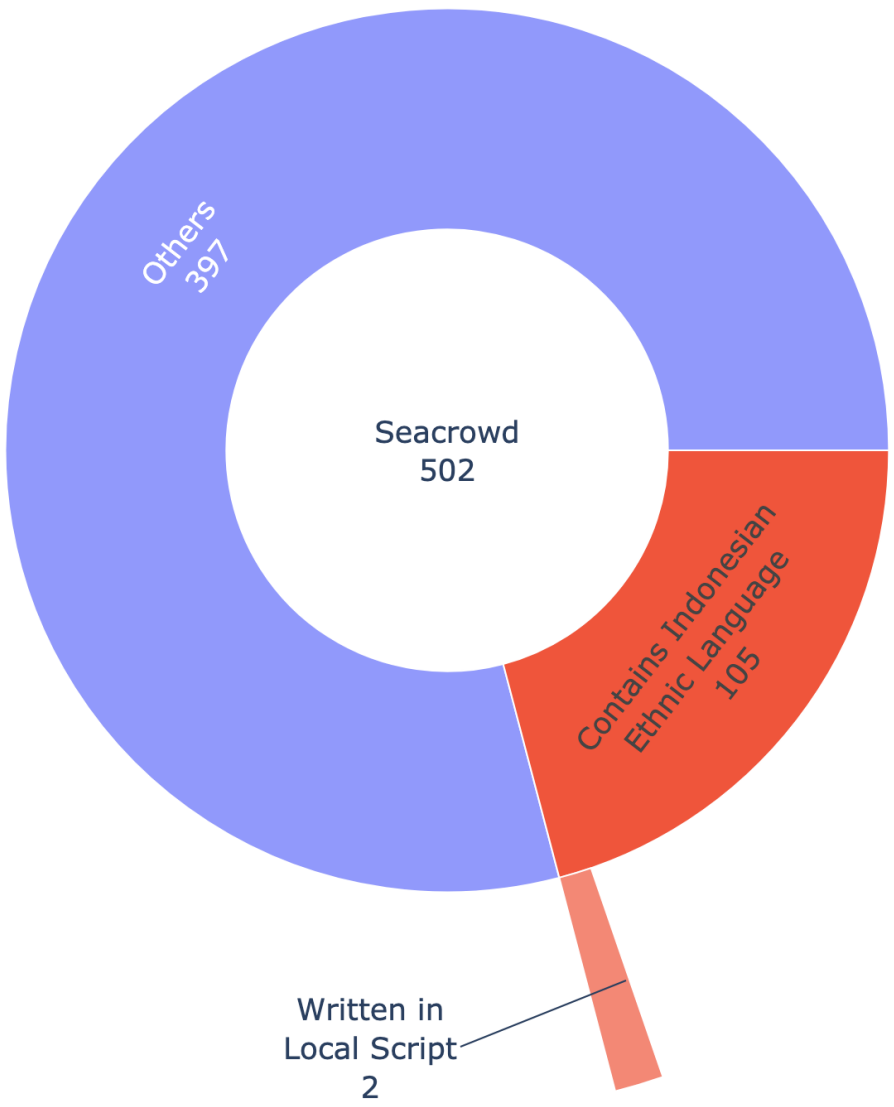

Figure 4: From the SEACrowd which contains 502 accepted datasets, 105 of them contains at least one of the 17 local Indonesian ethnic languages (lam, lpj, abl, ace, zsm, jav, xdy, bug, mak, sun, mad, bjn, bbc, btk, btx, min, ban) and only two of them are written in the original script.

## E Data Creation

In this section, we provide documentation of our data collection process. Figure 6 illustrates our manual process of unbinding books before scanning the text. We then annotate and train a segmentation method, as shown in Figure 7, as our first step. The statistics of the data used for image segmentation finetuning are shown in table 9. Next, we proceed with the annotation process to correct the segmentation, apply OCR, transliterate, and translate our data using LabelStudio. The annotation interface is shown in Figure 5.

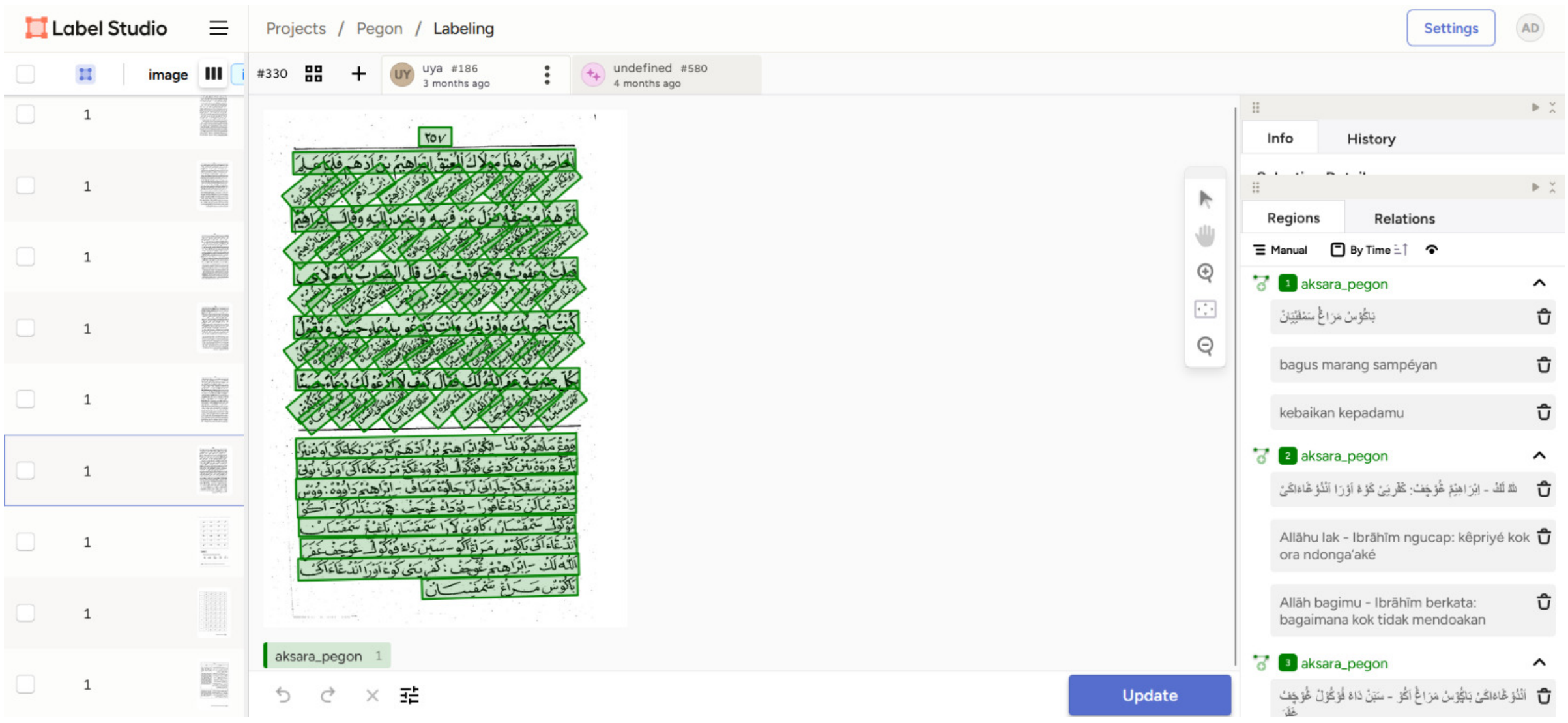

Figure 5: LabelStudio interface for annotation

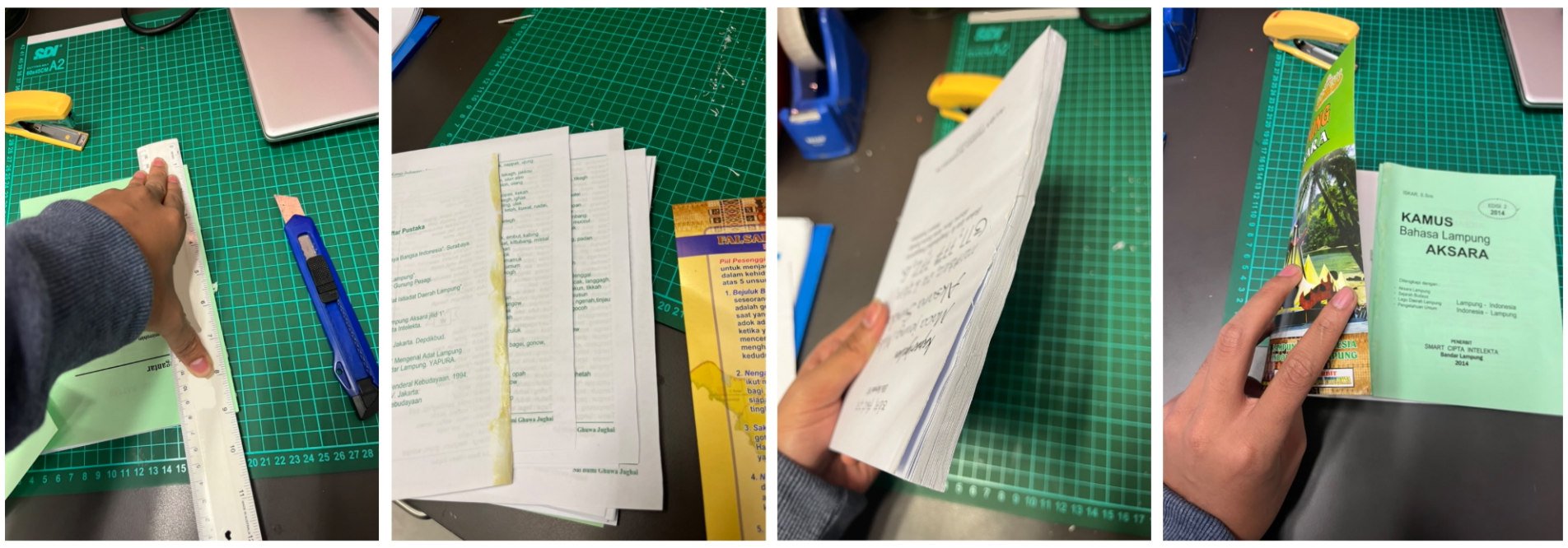

Figure 6: The process of unbinding resource books using simple tools such as cutter and ruler.

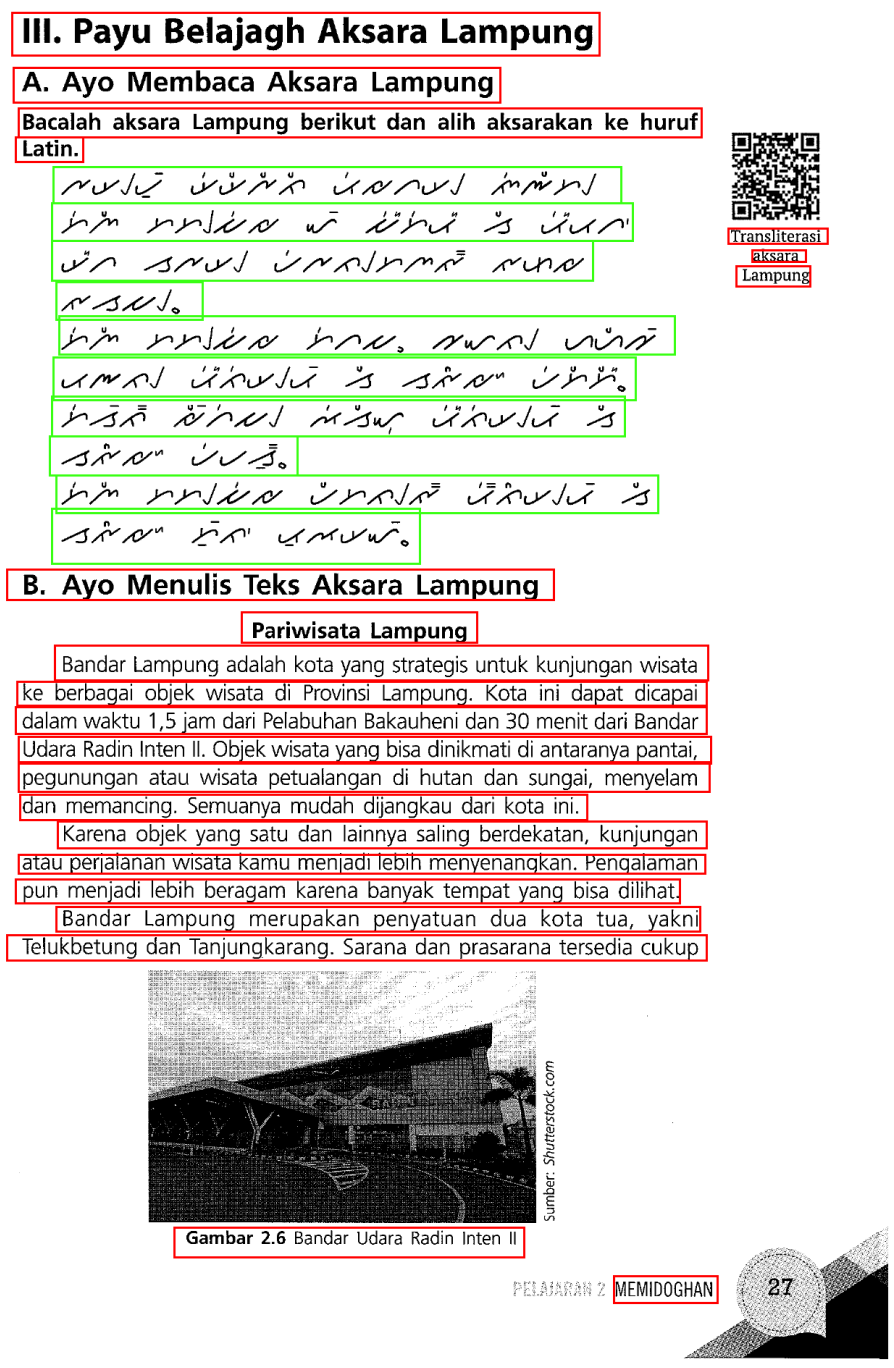

III. Payu Belajagh Aksara Lampung

A. Ayo Membaca Aksara Lampung

Bacalah aksara Lampung berikut dan alih aksarakan ke huruf Latin.

Transliterasi aksara Lampung

B. Ayo Menulis Teks Aksara Lampung

Pariwisata Lampung

Bandar Lampung adalah kota yang strategis untuk kunjungan wisata ke berbagai objek wisata di Provinsi Lampung. Kota ini dapat dicapai dalam waktu 1,5 jam dari Pelabuhan Bakauheni dan 30 menit dari Bandar Udara Radin Inten II. Objek wisata yang bisa dinikmati di antaranya pantai, pegunungan atau wisata petualangan di hutan dan sungai, menyelam dan memancing. Semuanya mudah dijangkau dari kota ini.

Karena objek yang satu dan lainnya saling berdekatan, kunjungan atau perjalanan wisata kamu menjadi lebih menyenangkan. Pengalaman pun menjadi lebih beragam karena banyak tempat yang bisa dilihat.

Bandar Lampung merupakan penyatuan dua kota tua, yakni Telukbetung dan Tanjungkarang. Sarana dan prasarana tersedia cukup

Sumber: Shutterstock.com

Gambar 2.6 Bandar Udara Radin Inten II

PELAJARAN 2 MEMIDOGHAN 27

Figure 7: Example of image segmentation annotation results that differentiate the alphabet text (red) with Lampung scripts (green)

| Scripts | #pages |
|---|---|
| Bali | 148 |
| Sunda | 138 |
| Lontara | 125 |
| Batak | 102 |
| Pegon | 101 |
| Jawa | 100 |
| Lampung | 100 |
| Jawi | 100 |

Table 9: Number of page annotated per local scripts for image segmentation tasks. Notes that some of the scripts have more than 100 pages of annotation since writers had partially annotated it.

## F Dealing with Lampung Script

Since Lampung script is not supported by Unicode, we have to use a custom font built by the local community to enable the annotators to write the text[8]. However, the text can only be read if the font is used, otherwise it will be nonsensical text. For example 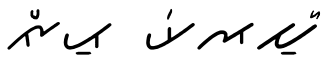 has to be written as “aibu mEGtuR” in Unicode which does not mean anything.

## G Supported Languages in LID

Typical LID does not support all languages covered in our dataset. The following are the languages they support.

| | ban | btx | jav | zsm | lpj | bug | sun |
|---|---|---|---|---|---|---|---|
| Langid | ✓ | ✗ | ✓ | ✓ | ✗ | ✗ | ✗ |
| LangDetect | ✗ | ✗ | ✗ | ✗ | ✗ | ✗ | ✗ |
| Fasttext | ✗ | ✗ | ✓ | ✓ | ✗ | ✗ | ✓ |
| CLD2 | ✓ | ✗ | ✓ | ✓ | ✗ | ✗ | ✗ |
| CLD3 | ✗ | ✗ | ✓ | ✓ | ✗ | ✗ | ✓ |

Table 10: Supported Languages across different language detection tools.

## H Data Validation

The following Table 11 shows the annotator’s agreement during our validation.

| **Scripts** | **Transcription** | | **Transliteration** | | **Translation** | |
|---|---|---|---|---|---|---|
| | **CER** | **WER** | **CER** | **WER** | **BLEU** | **chrf++** |
| Lampung | 0.008 | 0.036 | 0.010 | 0.033 | 98.350 | 99.207 |
| Jawi | 0.003 | 0.003 | 0.002 | 0.006 | 97.653 | 98.788 |
| Bali | 0.001 | 0.012 | 0.002 | 0.007 | 95.631 | 96.588 |
| Batak | 0.008 | 0.057 | 0.004 | 0.010 | 96.212 | 97.265 |
| Jawa | 0.054 | 0.544 | 0.010 | 0.031 | 93.103 | 95.574 |
| Lontara | 0.048 | 0.121 | 0.062 | 0.214 | 48.926 | 66.068 |
| Pegon | 0.013 | 0.047 | 0.009 | 0.021 | 93.861 | 96.202 |
| Sunda | 0.008 | 0.011 | 0.005 | 0.007 | 98.190 | 96.682 |

Table 11: Annotator-validator agreement across tasks: evaluating the quality of transcription, transliteration, and translation in the data validation process.

[8] https://aksaradinusantara.com/fonta/font/Kaganga_21key=9e4d311c4c09970827bca94ab8d6fe1c

# I Annotation Guideline

The following is the guideline we provide to annotators. The instructions and video tutorial are given in Indonesian, as it is the language they are fluent in, whereas not everyone may be familiar with English.

## Annotation Guideline: Transkripsi, Transliterasi, dan Translasi Aksara Daerah

### Tugas Utama

1. **Transkripsi** gambar menjadi aksara daerah
2. **Transliterasi** aksara daerah ke tulisan latin dalam bahasa daerah
3. **Translasi** bahasa daerah dalam latin ke bahasa indonesia

Tonton Video penjelasan ini:
https://youtu.be/<redacted>

Perhatikan:
1. Harus 4 titik polygon
2. Perbaiki bounding-box jika ada yang salah

### Langkah Pengerjaan

1. Akses Annotation Platform
    - Buka folder pada label studio yang telah diinstall sesuai dengan aksara daerah yang dipilih.

2. Proses Setiap Gambar dalam Folder
    - **Transkripsi:**
      Lakukan **transkripsi** gambar menjadi tulisan ketik menggunakan aksara daerah dengan cara mengklik bounding box aksara daerah dan mengisi form yang muncul untuk transkripsi.

      **Contoh:**

      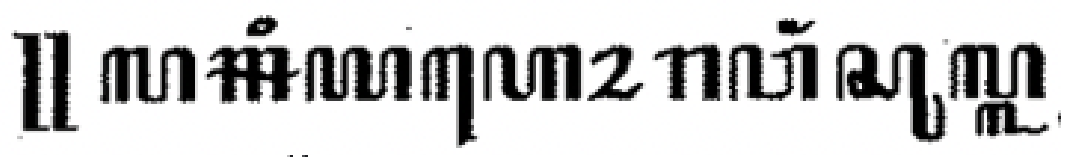

      Transkripsi: 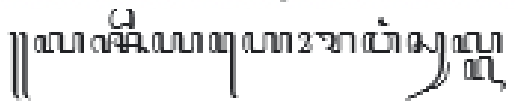

    - **Transliterasi:**
      Lakukan **transliterasi** aksara daerah yang telah dikerjakan pada tahap transkripsi menjadi tulisan latin dengan cara mengklik bounding box gambar aksara daerah dan mengisi form yang muncul untuk transliterasi.

      **Contoh:**

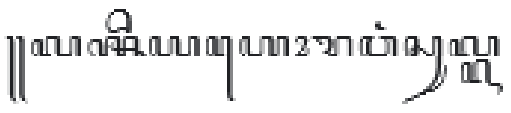

Transliterasi: “Lasiya ora wangsulan”

- **Translasi:**

  Lakukan **translasi** tulisan latin bahasa daerah yang telah dikerjakan pada tahap transliterasi menjadi Bahasa Indonesia dengan cara mengklik bounding box pada gambar aksara daerah dan mengisi form yang muncul untuk translasi.

  **Contoh:**

  “Lasiya ora wangsulan”

  Translasi: Lasiya tidak pulang

4. Catatan Tambahan
   - Pastikan setiap langkah diikuti dengan cermat untuk menjaga akurasi dan konsistensi.
   - Jika ada ketidaksesuaian antara gambar dan teks yang ditranskripsi, beri catatan pada platform anotasi untuk koreksi lebih lanjut.
   - Periksa kembali setiap transkripsi, transliterasi, dan translasi untuk memastikan keakuratannya.
   - Pastikan penggunaan aksara daerah dan bahasa daerah sesuai dengan kaidah yang berlaku.

## J Model Output Examples

In this section, we show some outputs produced by the models to highlight their errors.

**Script Hallucination** Models struggle extremely with transcription tasks. Most zero-shot OCR models hallucinate by writing in a completely different script, such as the Devanagari script; or giving insensible output such as repeated strings. Figure 8 showcases that gemini even hallucinates Indonesian local scripts with Hindi.

**Correct Script but Incorrect Text** In some cases, for example, fine-tuned OCR models or high-performing LLMs such as GPT, can recognize the script and write in the correct script, but the text may be incorrect.

**Text Hallucination** This is perhaps one of the most common cases. When the model is asked to translate to Indonesian, it may hallucinate an entirely unrelated meaning of the text. Similarly, when asked for transliteration, the model may hallucinate the transliteration, or just repeating the input again.

---

**Input**: ᨆᨛᨔᨑᨛᨀᨔᨗᨂᨙ
**Gold Transliteration**: Masero kessingnge
**Gold Translation**: Yang indah (*The beautiful one*)
**Transliteration Prediction**: Nanggala1 ᨆᨛᨔᨑᨛᨀᨔᨗᨂᨙ Transliteration: Nanggala-1 ᨆᨛᨔᨑᨛᨀᨔᨗᨂᨙ Transliteration: Nanggala-1 ᨆᨛᨔᨑᨛᨀᨔᨗᨂᨙ
**Translation Prediction**: Masero kessingnge (masero kessingnge) adalah sebuah lagu yang dibuat oleh band rock Indonesia, Gigi. Lagu ini dirilis pada tahun 2007 dan menjadi salah satu lagu hits dari album mereka, "Masero Kessingnge." Lagu ini memiliki lirik yang unik dan menarik, serta memiliki irama yang khas. Lagu ini juga memiliki video musik yang menarik dan menampilkan adegan-adegan yang menarik. Lagu ini menjadi sangat populer di kalangan penggemar musik rock di Indonesia

---

Table 12: Hallucinaton example for translation on Llama 3.1

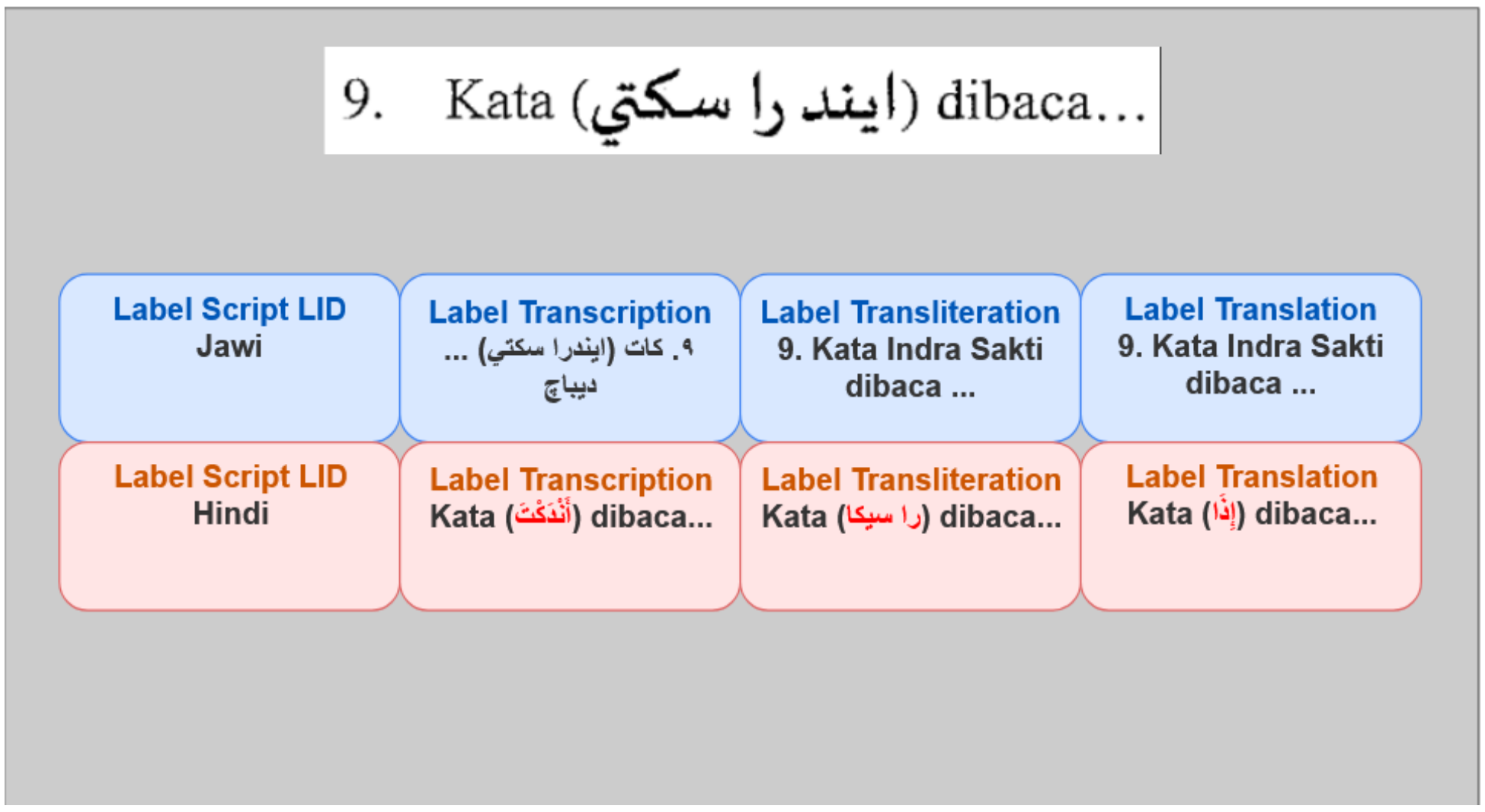

Figure 8: Example case of Gemini Hallucinating on multiple task.

## K Experiment Configuration

Table 13 lists all models that we used in this paper to their respective checkpoint. Image segmentation and OCR are not typically zero-shot settings. Therefore some of the baselines are finetuned models using PaddleOCR-based framework, such as PP-OCRv3 and DBResNet-50. For the image segmentation tasks, we finetune PP-OCRv3 (detection) and DBResNet-50 for each script using labeled data from bounding-boxes annotators with 9:1 train-validation split. The finetuned models were tested on separate data annotated by natives. The test data was also used to benchmark zero-shot capability of SAM-ViT model. On the OCR task, we benchmark the PP-OCRv3 by performing 5-fold validation as our data was not split for training. In each fold, we use the training set to finetune and the test set to evaluate the model, then, we provide the average fold results.

| Model | Checkpoint/URL |
|---|---|
| PP-OCRv3 (Detection) | ch_PP-OCRv3_det_student |
| PP-OCRv3 (Recognition) | ch_PP-OCRv3_rec_distillation |
| SAM-ViT | facebook/sam-vit-base |
| DBResNet-50 | DBResNet-50_vd |
| Intern-VL | InternVL2_5-8B |
| LLaVA-NeXT | LLaVA-v1.6-mistral-7B-hf |
| Llama 3.2 | Llama3.2-11B-Vision |
| GPT-4o | GPT-4o-2024-08-06 |
| Gemini Flash | gemini-1.5-flash |
| Cendol | Cendol-7b-llama2-7b-inst |
| Sailor-7B | Sailor-7B |
| Bloomz-7B1 | Bloomz-7B1 |
| Aya-23-8B | aya-23-8B |
| Llama-3.1-8B | Llama-3.1-8B |
| NLLB-3.3B | NLLB-3.3B |
| LangID | LangID |
| FastText | Fasttext |
| CLD2 | CLD2 |
| CLD3 | CLD3 |
| Franc | Franc |

Table 13: Models used in this work.

# L Full Result

In this part, we provide results across all tasks on various metrices.

| Model | Sunda | Pegon | Lontara | Jawi | Jawa | Batak | Bali | Lampung |
|---|---|---|---|---|---|---|---|---|
| **Transliteration from Image** | | | | | | | | |
| InternVL2.5-8B | >1 | >1 | >1 | >1 | >1 | >1 | >1 | >1 |
| LlaVA-v1.6-7B | >1 | >1 | >1 | >1 | >1 | >1 | >1 | >1 |
| Llama3.2-11B | >1 | >1 | >1 | >1 | >1 | >1 | >1 | >1 |
| GPT-4o | >1 | >1 | >1 | .95 | >1 | >1 | >1 | >1 |
| Gemini Flash | >1 | >1 | >1 | >1 | >1 | >1 | >1 | >1 |
| **Transliteration from Local Aksara** | | | | | | | | |
| Cendol-7b | >1 | >1 | >1 | >1 | >1 | >1 | >1 | - |
| Sailor-7B | >1 | >1 | >1 | >1 | >1 | >1 | >1 | - |
| Bloomz-7B1 | >1 | >1 | >1 | >1 | .99 | >1 | >1 | - |
| Aya-23-8B | >1 | >1 | >1 | >1 | >1 | >1 | >1 | - |
| Llama-3.1-8B | >1 | >1 | >1 | >1 | >1 | >1 | >1 | - |
| Llama-3.2-11B | >1 | >1 | >1 | >1 | >1 | >1 | >1 | - |
| InternVL2.5-8B | >1 | >1 | >1 | >1 | >1 | >1 | >1 | - |
| GPT-4o | .57 | >1 | .92 | .60 | .87 | >1 | .97 | - |
| Gemini Flash | >1 | .98 | >1 | .78 | .88 | >1 | >1 | - |

Table 14: Word Error Rate (WER) comparison across models for image-based and aksara-based transliteration.

| Model | ban | btx | jav$_{jj}$ | zsm | bug | jav$_{jp}$ | sun |
|---|---|---|---|---|---|---|---|
| **Translation from Image** | | | | | | | |
| GPT-4o | 0.00 | 0.10 | 0.00 | 10.40 | 0.00 | 0.37 | 0.84 |
| Gemini Flash | 0.00 | 0.00 | 0.00 | 1.29 | 0.00 | 0.00 | 0.06 |
| InternVL2.5-8B | 14.06 | 4.50 | 9.12 | 12.01 | 8.93 | 7.51 | 7.83 |
| LLaVA-v1.6-7B | 8.35 | 1.34 | 5.34 | 4.28 | 4.72 | 3.88 | 3.74 |
| Llama3.2-11B | 4.81 | 1.18 | 2.91 | 4.44 | 3.14 | 2.62 | 2.92 |
| **Translation from Local Aksara** | | | | | | | |
| Cendol | 11.65 | 5.31 | 11.25 | 13.18 | 12.29 | 9.58 | 11.25 |
| Sailor-7B | 7.02 | 2.23 | 6.26 | 12.01 | 4.96 | 4.20 | 4.83 |
| bloomz-7b1 | 11.12 | 10.08 | 12.29 | 12.30 | 13.40 | 7.24 | 11.41 |
| aya-23-8B | 4.77 | 4.04 | 5.48 | 13.90 | 7.49 | 4.03 | 6.62 |
| Llama-3.1-8B | 12.41 | 7.47 | 9.71 | 19.67 | 13.29 | 5.25 | 9.51 |
| GPT-4o | 0.00 | 0.00 | 2.36 | 18.11 | 0.45 | 3.88 | 3.18 |
| Gemini | 0.00 | 0.00 | 1.73 | 13.53 | 1.32 | 3.63 | 1.73 |
| NLLB-3.3B | 2.85 | 2.31 | 3.57 | 20.83 | 9.31 | 5.23 | 6.86 |
| InternVL2.5-8B | 0.00 | 0.00 | 0.00 | 0.00 | 0.00 | 0.00 | 0.00 |
| **Translation from Romanized Script** | | | | | | | |
| Cendol | 19.10 | 27.91 | 35.63 | 43.36 | 16.77 | 28.80 | 34.46 |
| Sailor-7B | 13.99 | 32.13 | 23.48 | 41.93 | 16.14 | 20.15 | 23.80 |
| bloomz-7b1 | 13.82 | 22.36 | 18.15 | 39.81 | 14.03 | 16.38 | 19.15 |
| aya-23-8B | 13.95 | 29.69 | 23.15 | 42.49 | 14.93 | 19.47 | 23.94 |
| Llama-3.1-8B | 15.55 | 28.65 | 23.12 | 39.57 | 16.22 | 26.09 | 25.50 |
| GPT-4o | 4.77 | 6.28 | 20.76 | 28.27 | 3.52 | 19.37 | 23.57 |
| Gemini | 1.96 | 0.75 | 8.27 | 10.76 | 1.02 | 1.20 | 8.88 |
| NLLB-3.3B | 20.19 | 32.05 | 33.73 | 48.89 | 24.07 | 31.37 | 36.47 |
| InternVL2.5-8B | 0.00 | 0.00 | 0.00 | 0.02 | 0.00 | 0.00 | 0.00 |

Table 15: BLEU performance on various languages for different translation tasks.